\documentclass[twocol]{ametsocV6.1}

\usepackage{comment}
\usepackage{hyperref}
\usepackage{cleveref}

\title{Lessons Learned \\ Reproducibility, Replicability, and When to Stop}

\authors{Milton S. Gomez\aff{a, b}, Tom Beucler\aff{a, b}, \correspondingauthor{Milton S. Gomez, milton.gomez@unil.ch}}

\affiliation{\aff{a}{Faculty of Geosciences and Environment, University of Lausanne, Switzerland}
\\\aff{b}{Expertise Center for Climate Extremes, University of Lausanne, Switzerland}
}

\abstract{
While extensive guidance exists for ensuring the reproducibility of one's own study, there is little discussion regarding the reproduction and replication of external studies within one's own research. To initiate this discussion, drawing lessons from our experience reproducing an operational product for predicting tropical cyclogenesis, we present a two-dimensional framework to offer guidance on reproduction and replication. Our framework, representing model fitting on one axis and its use in inference on the other, builds upon three key aspects: the dataset, the metrics, and the model itself. By assessing the trajectories of our studies on this 2D plane, we can better inform the claims made using our research. Additionally, we use this framework to contextualize the utility of benchmark datasets in the atmospheric sciences. Our two-dimensional framework provides a tool for researchers, especially early career researchers, to incorporate prior work in their own research and to inform the claims they can make in this context. 
}

\begin{document}

\newcounter{saveenumerate}
\makeatletter
\newcommand{\enumeratext}[1]{%
\setcounter{saveenumerate}{\value{enum\romannumeral\the\@enumdepth}}
\end{enumerate}
#1
\begin{enumerate}
\setcounter{enum\romannumeral\the\@enumdepth}{\value{saveenumerate}}%
}
\makeatother

\maketitle

%
%
%
%
%
%

%

\section{Introduction}
\label{sec:intro}

\defcitealias{national2019reproducibility}{NASEM, 2019}\
\defcitealias{mullendore}{Mullendore et al., 2020}
\defcitealias{GRRIEn}{Carter et al., 2023}
\defcitealias{TCFP}{NOAA and CIRA, 2010}\

Should scientists focus their efforts on reproducing previous studies? The answer to this question seems straightforward---the reproduction of scientific studies has been a recurrent focus over concerns that published works may be irreproducible or irreplicable \citep{nichols2021better, hoffmann2021multiplicity}. Additionally, reproducibility of results is taken as evidence of scientific robustness across numerous domains \citep{gervais2021practical, ahadi2016replication, baker20161}. In the atmospheric sciences, this is evidenced by data replication efforts (e.g., the ESGF Data Replication Project---\citep{cinquini2014earth}) and benchmark dataset requirements (e.g., \textit{``The availability of an existing solution [...] will also enable the easy reproduction of published results to verify that the workflow is consistent.''}, \citep{dueben2022challenges}). However, the terms reproducibility and replicability themselves can be a source of confusion---the (US) National Academies note that depending on the field, what is meant by each of these two terms can be ``inconsistent, or even contradictory'' \citepalias{national2019reproducibility}, a fact that is echoed by others in the literature \citep{devezer2021case}. In light of the substantial efforts by the academies in guiding scientists' endeavors to improve the reproducibility and replicability of their own works, let us borrow their definitions---\mbox{1) Reproducibility} entails obtaining consistent computational results using the same input data, computational steps, methods, code, and conditions of analysis; and 2) Replication entails obtaining consistent results across studies, aimed at answering the same scientific question, each of which has obtained its own data. Still, we note that reproduction and replication of work does not itself provide proof of a given claim \citep{baker20161, goodman2016does}, but each plays an important role in science, and efforts towards them are laudable.

\defcitealias{Stall_2023}{Stell et al., 2023}
This begets the question of how to address reproducibility and replication in scientific works. While we can point to sources in the literature providing guidance in making researchers' findings more reproducible \citepalias[e.g.,][]{national2019reproducibility, mullendore, GRRIEn, Stall_2023}, we found limited guidance regarding the reproduction and replication of others' work in one's own research studies---motivating our study.

\section{Lessons Learned from Predicting Tropical Cyclogenesis}
\label{sec:case}

Reproduction and replication are important, and I propose that we keep this in mind as we take a look at our case study, in which we aimed to apply modern machine learning techniques to a well-known problem in meteorology---the prediction of tropical cyclone formation, or ``tropical cyclogenesis'' (see \textit{S1:Scientific Motivation}). We believed that interpreting such models could provide scientific insight into an aspect of tropical cyclones that has been described as both \textit{vexing} and \textit{intractable} \citep{emanuel2018100}. We wanted to select a well-established method as a baseline---assessing new models against well-established methods is, after all, generally the main interest of a study \citep{jolliffe2005comments}---and we thus selected the base study \citep{schumacher2009objective} for the 2014 Tropical Cyclone Formation Probability product \citepalias{TCFP}. Taking inspiration from \cite{devezer2021case}, we set up a number of definitions detailed in \Cref{app:A1}. Armed with these definitions, we begin by saying we want to develop a model $M$ through a study $\xi$ that depends in part on the inputs ($\mathbf{X}$) and model $M_{\text{\tiny BASE}}$ in the base \mbox{study, i.e.}:
\begin{align}
 \xi \sim \left( {M,
                     M_{\text{\tiny BASE}},
                     \mathbf{X}
                     } \right).   
\end{align}
While we sampled our input variables from ECMWF's European Reanalysis 5 \citep{81046} (ERA5 hereafter), we were aware that $M_{\text{\tiny BASE}}$ was not developed using $X$. After all, ERA5 was not available when $M_{\text{\tiny BASE}}$ was first developed, and $M_{\text{\tiny BASE}}$ instead relied on a combination of GFS analysis/reanalysis data and satellite data---henceforth denoted as $X_{BASE}$ \citep{schumacher2009objective}. Furthermore, we knew that simply transferring the model---assuming all of the variables in $X_{BASE}$ used to fit $M_{\text{\tiny BASE}}$ were available in $X$---was not a straightforward task. This is evidenced by the rich literature on the subject of dataset shifts and generalization, including, e.g., \citet{wang2022generalizing, malinin2021shifts, quinonero2008dataset}. As such, we needed to develop a version of $M_{\text{\tiny BASE}}$ that was trained using samples taken from $X$---\mbox{i.e., $M_{\text{\tiny BASE.ERA5}}$}---and we reasoned that the best way to do so was to first faithfully reproduce $M_{\text{\tiny BASE}}$ (\mbox{i.e., the} hyperparameters $ \mathbf{\theta_{hyper}}$ and trainable parameters $\mathbf{\theta_{train}}$ associated with $M_{\text{\tiny BASE}}$) to ensure that our understanding of the model was sufficient to train $M_{\text{\tiny BASE.ERA5}}$. 

Despite the help of the authors of \cite{schumacher2009objective}, our attempts at reproducing the base study would undergo several complications. First, the model that was shared with us had been updated since the publication of \cite{schumacher2009objective}---thus, this reference would be insufficient for our efforts. Though there was an AMS presentation highlighting part of the updates to the model \citep{schumacher2014update} and the authors provided the original developmental dataset (\mbox{i.e., $X_{BASE}$}), the model was developed for operational use rather than publication and a number of developmental details were unavailable. In a sense, we were working off an incomplete set of schematics; noting that ignorance of model training choices is detrimental \citep{EDSMcGov}. As a result, it would take numerous months to develop a working understanding of the model as it existed at the time (model details are given in \textit{S2:Base Model Details}). We had access to performance metrics of the post-processed model (e.g., the Brier score, ROC curve, and reliability), but we were missing the fitting metric of the sub-model for classification (the confusion matrix of the LDA).
Yet another obstacle we faced was one that is more frequently encountered---the original Linear Discriminant Analysis (LDA) model was developed with then commonly used Fortran libraries, while our efforts utilized Python equivalents (e.g., Scikit-Learn---\citealt{scikit-learn}). Subtle implementation differences (e.g., in the solvers used), lying in codebases too time-consuming to compare in depth, made it difficult to achieve the same LDA coefficients; noting here that since I lacked the original sub-model performance metrics, the coefficients were the target for my model reproduction efforts. We could be said to be trying to reproduce the model using a potentially different set of tools. 
And still we added another layer of complexity to our efforts ---we wanted to find a way to improve the model. After all, we wanted to give the LDA approach the best shot in our comparisons---we believe that baselines should be made robust to produce fair comparisons (details on attempted improvements are given in \textit{S3:Proposed Improvements}). We were simultaneously trying to develop the capacity to emulate our predecessors while also increasing the height of the bar. In summary: we were attempting to hit a blurry target with a model constructed using incomplete schematics and a potentially different set of tools, all while attempting to find ways to improve the model.

With our efforts described as such, it may be no surprise that they were not fruitful. While we could use the model to reproduce the images on the product website, training our own models resulted in algorithms that exhibited a wide variance in metrics depending on the data split. The model with our own attempts at improvement seemed to provide more stable results, but comparisons were within the standard deviations of the metrics as measured by our cross validation scheme. Our progress became so slow that we decided it no longer made sense to continue with these efforts, at which point we started to analyze where the project had gone awry. This led us to a strange realization: while there is plenty of literature offering guidance in ways to make your own work more reproducible and replicable, we were not aware of similar documents describing how to gauge the level of reproducibility and replicability of others' works serving as the baseline in your study. This raises essential questions: How much effort should be allocated to replicability efforts? And how can one effectively guide and quantify replicability efforts to ensure rigor in the study's conclusions?


\section{A Short Guide to Reproduction and Replication}
\label{sec:guide}

The level of reproduction and replication of previous work can have a significant impact on the scientific claims that can be made in light of said work. Thus, the aim of this work is to address a perceived lack of guidance with regard to gauging the extent of reproduction/replication of others' work achieved in one's own studies.
Furthermore, we expect that methods for evaluating reproduction and replication can provide guidance in evaluating future projects, and potentially serve as an evaluation tool during review processes. 

\begin{figure*}
    \centering
    \includegraphics[width=\textwidth]{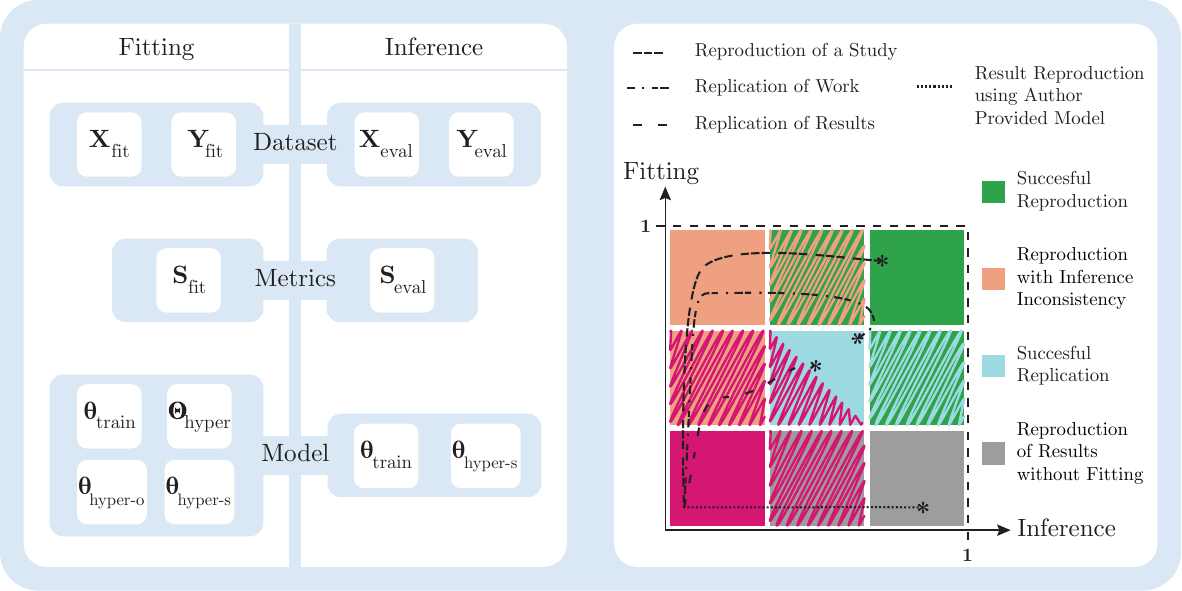}
    \caption{We introduce a two-component framework for assessing replication, employing distinct axes to represent the fitting and inference aspects. The y-axis (``Fitting'') evaluates the concordance in dataset, metrics, and parameters utilized for model \textit{training}, while the x-axis (``Inference'') assesses the alignment in dataset, metrics, and parameters employed for model \textit{evaluation}. Empirical scores guide the positioning along each axis, with the depicted paths on the right illustrating typical trajectories in research endeavors. $\bm{\theta}_{hyper-o}$ denotes the optimization hyperparameters, $\bm{\theta}_{hyper-s}$ denotes the structural hyperparameters (e.g., the depth, width, and type of a neural network's layers), and $\bm{\Theta}_{hyper}$ denotes the searchable hyperparameter space. Other Variables are defined in the appendix.}
    \label{fig:axes}
\end{figure*}



For simplicity's sake, we aim to quantify reproduction and replication using two axes, as depicted in \Cref{fig:axes}. The vertical axis represents the ability to \textit{fit} a base model (e.g., the model described by \cite{schumacher2009objective, schumacher2014update})---measured by one's confidence in the reproduction of the fitting dataset, metrics, and model associated with the base study---represented by a value ranging from 0--1. In contrast, the horizontal axis represents the ability to use the model in \textit{inference} mode to evaluate its performance---measured by one's confidence in the reproduction of the evaluation dataset, metrics, and model---once again represented by a value ranging from 0--1. The definition of these axes provides a useful analytical tool, but we still lack a method for evaluating our position on the axes.

First, what precisely defines confidence in the equivalence of our datasets, metrics, and models can vary depending on our understanding of the research subject and the particularities of the question being addressed. In early research, for example, agreement on the sign of the coefficients associated with covariates in a linear model can already be ambitious. In light of this variance, ways to determine our position on these axes are not immediately apparent and thus evaluation of our positioning on these axes is something that must be considered per project. Still, in our hope to provide guidance to researchers we propose that one's position along these axes be evaluated empirically using the questionnaire made available at \href{https://github.com/msgomez06/RepoRepli}{github.com/msgomez06/RepoRepli}---whose questions can be found in \textit{S4:Questionnaire}.

In \Cref{fig:axes} we further propose a set of curves describing what we consider are typical trajectories taken by researchers when undertaking research. Noting that the trajectories start in the red, bottom-left corner of the diagram, which represents a point in which the researcher is not yet familiar with the study and is not yet in possession of the pertinent datasets. For reproduction and replication studies, the trajectory continues with increasing confidence in the ability to produce an equivalent set of input and target samples (i.e., $\mathbf{X}_{fit}$ \& $\mathbf{Y}_{fit}$), an equivalent set of fitting metrics $\mathbf{S}_{fit}$ (e.g., accuracy, recall, root mean square error), and thereby fit the model defined by a set of equivalent model hyperparameters $\bm{\theta}_{hyper}$, resulting in a set of trainable parameters $\bm{\theta}_{train}$ equivalent to those in the base study. We note that at this stage the path is in the orange, top-left corner of the diagram, where if the researcher would be unable to develop confidence in the use of the model in inference mode for evaluation purposes, this would be indicative of inconsistencies in the evaluation of the model in inference mode. Assuming the researcher is instead able to use the model in inference mode and develop confidence that the evaluation dataset and protocols are equivalent to those in the base study, their trajectory should move to the top-right corner, indicative of a successful reproduction procedure. If the aim of the research project is replication (e.g., to establish a framework's robustness), the researcher moves from a regime in which they are confident that all aspects of the fitting and inference evaluation are equivalent, and makes modifications that decrease this degree of confidence. This includes, e.g.: changing the fitting and evaluation datasets by adding or removing covariates; changing the model hyperparameters used to make predictions; and changing the evaluation protocol to focus on different aspects of the predicted outcomes. We emphasize that these are idealized examples of trajectories - actual research trajectories will vary as projects progress.

Second, we propose that this two-axis framework (and the trajectory taken through it) can help inform the type of claims that can be made from one's research. Let's consider the AI-model baselines described by \citep{wb2}, which are described as being trained on datasets covering a different number of years. While it is understandable that the models were not retrained for comparison---model training times ranged from 5.5 days on a single Nvidia A100 GPU to 4 weeks on 32 TPU v4 devices---this mismatch in training data results in a comparison between models that is less fair. This is pointed out by \cite{wb2}, who note that training on additional data results in improved performance. Figure 5 in \cite{CVStrat} also demonstrates how a model's skill can vary significantly depending on which year is used for evaluation. In our framework, we see that this relates to less confidence in $\mathbf{X}_{fit}$ being equivalent, independently of the inference dataset used. Furthermore, without a quantification of the uncertainty in $\mathbf{S}_{fit}$ and $\mathbf{S}_{eval}$, it becomes more difficult to compare the scores achieved by the models. 
Still, by providing a consistent evaluation framework including the dataset and metrics, \cite{wb2} facilitate reproduction and replication studies.

\begin{figure*}
    \centering
    \includegraphics[width=\textwidth]{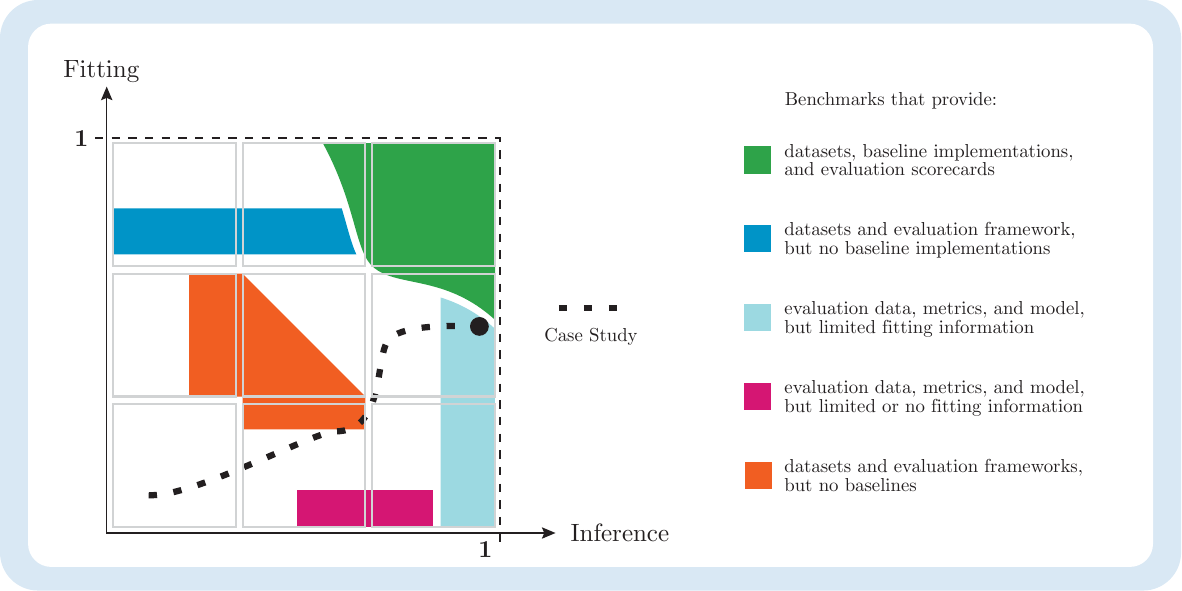}
    \caption{Benchmark datasets that provide baseline implementations and evaluation scorecards (green) allow users to start in the ``successful reproduction'' area. The absence of baseline implementation (dark blue) limits inference, while little (light blue) to no (pink) fitting information (light blue) prevents re-fitting of the model. The absence of baseline models (orange) prevents objective comparison to prior work. We note that benchmark datasets providing baselines. Note that the green section describes benchmark datasets fulfilling the first-order requirements and requirement 5 of the second-order requirements described by \cite{dueben2022challenges}, while the orange section describes benchmark datasets fulfilling only the first-order requirements. This difference is due to the inclusion of trainable baselines for verifying workflows. }
    \label{fig:benchmarks}
\end{figure*}

We can further use the framework to describe the advantages provided by benchmark datasets more generally, as shown in \Cref{fig:benchmarks}. We start by considering the first-order and second-order requirements for benchmark datasets described by \cite{dueben2022challenges}. The first four requirements focus on data access, problem definition, and evaluation metrics. The confidence provided by benchmark datasets thus allow researchers to start their trajectories in an advantageous position---the researchers position on the axes should be high considering the dataset confidence and metric confidence stemming from the use of a benchmark. Similarly, requirement 5 focuses on providing baselines. Quoting \cite{dueben2022challenges}: "\textit{The availability of an existing solution will facilitate the use of the dataset but will also enable the easy reproduction of published results to verify that the workflow is consistent.}" We see here that providing a baseline that allows the verification of a consistent workflow means that researchers are able to \textit{begin} in the top-right "successful reproduction" area of the diagram, greatly facilitating replication studies. With requirements 6-8, we note one of the limitations of the two-axes framework: while the representation allows us to determine the confidence in our ability to reproduce and replicate a study, it does not include a measure of the \textit{quality} of the dataset, evaluation protocol, and model. \cite{dueben2022challenges} define in these requirements for visualizations, diagnostics, physical consistency, and computational performance - all aspects that are independent of the reproduction and replication of base studies but rather focus on the rigor of the scientific question being addressed or the computational efficiency of the model implementation. Still, the framework shows clear advantages associated with using benchmark datasets in reproduction and replication efforts.

Finally, we take this opportunity to circle back to the case study presented in \Cref{sec:case}. We were able to reproduce the resulting maps from the base study well but hit a limit with our confidence in improving our ability to fit the model---depicted by the case study path in \Cref{fig:benchmarks}. Looking back, perhaps our goals for confidence in the evaluation of the model were too high and unachievable in the context of trying to reproduce a production algorithm instead of the original study \citep[i.e.,][]{schumacher2009objective}. Alternatively, we could have decided to not pursue a replication study and simply use the study as inspiration for the baseline while being clear that there were fundamental differences in our implementation versus the one used in production, limiting the claims we could make. Regardless, we believe that the limitations encountered in our case study could help other researchers in similar circumstances with their decision-making.

\section{Conclusions}
\label{sec:conclusions}

In conclusion, while making a study reproducible can be informed by checklists (\mbox{e.g., those} provided by \cite{ESIP2022,dueben2022challenges}), we emphasize the need for a holistic approach when replicating an already-completed study. As a first step, we propose the use of a questionnaire that allows researchers to place themselves on the proposed two-dimensional representation of a study. The evaluation of our ability to reproduce and replicate the fitting of the model, as well as our ability to do the same for the evaluation of the model in inference mode, serves as a diagnosis to focus efforts for improvements to our approach at reproducing/replicating previous studies. Furthermore, carefully considering the aspects that we are able (or unable) to reproduce with regard to base studies can inform the types of claims we make after completing the study.

This underscores the utility of benchmark datasets in atmospheric science, which provide consistent datasets and evaluations to allow model comparison. This is especially important considering model sensibilities to spatiotemporal partitioning and the claims we can make as a result. By providing accessible datasets, metrics, and baseline models, study reproduction \& replication become streamlined and lead to better scientific discussions.  

\acknowledgments
We thank Andrea Schumacher for her assistance in helping us reproduce \citep{schumacher2014update} and for her feedback, which greatly improved the manuscript.  

%
%
\datastatement

The questionnaire and associated files described in the manuscript can be accessed in the following GitHub Repository: \hyperlink{https://github.com/msgomez06/RepoRepli}{https://github.com/msgomez06/RepoRepli}


%

\appendix
\label{app:A1}

In this appendix, we formally define studies, datasets, metrics, and models, which help us define the framework and empirical scores used in the main text. We begin by considering a measurable event that we are interested in. This can be cyclogenesis, surface precipitation, etc.
\begin{enumerate}
    \item\label{th:targets} There is a random variable $Y$ with discrete measurements $\bm{Y}_{obs}$, \\\mbox{$\mathbf{Y}_{obs} \in \{y_{event}, y_{non-event}\}$.}
    \enumeratext{We believe that there are variables associated with the event that can be useful to predict event occurrences, and thus we say:}
    \item\label{th:inputs} There is a random variable $\mathbf{X}$ with observed measurements $\mathbf{X}_{obs}$ that we believe can be used to predict $\bm{Y}_{obs}$
    \enumeratext{Having defined an event and an associated set of predictors, we want to find a model $M$ which uses a set of learnable parameters $\bm{\theta}_{learn}$ and the predictors $\mathbf{X}_{obs}$ to make a set of predictions $\mathbf{\hat{Y}}$ of the event.}
    \item\label{th:model}$\mathbf{\hat{Y}} = \bm{M}\left( \mathbf{X}_{obs}, \; \bm{\theta}_{learn} \right)$
    \enumeratext{Since parameters are learned from measurements, we obtain two or more independent samples of $\mathbf{X}$ and $\bm{Y}$ - a first time to obtain a model fitting set $\left\{ \mathbf{X}_{fit}, \mathbf{Y}_{fit} \right\}$ and a second time to obtain a model evaluation set $\left\{ \mathbf{X}_{eval}, \mathbf{Y}_{eval} \right\}$}
    \item{$\mathbf{X}_{fit}, \mathbf{X}_{eval}  \in \mathbf{X}, \quad \mathbf{Y}_{fit}, \mathbf{Y}_{eval}  \in \mathbf{Y}, \quad \text{with} \quad\mathbf{X}_{fit} \cap \mathbf{X}_{eval} = \emptyset$}
    \enumeratext{The model is defined by: 1) a set of structural hyperparameters $\bm{\theta}_{hyper-s}$ specifying the number of learnable parameters and the parameters are used to make the prediction (\mbox{e.g., the} depth, width, and type of layers in a neural network); 2) a method of fitting the model, specified by a set of optimization hyperparameters $\bm{\theta}_{hyper-o}$ controlling how learnable parameters should be adjusted during training (e.g., the batch size, learning rate, and optimization algorithm); and 3) the hyperparameter space $\bm{\Theta}_{hyper}$ considered when searching of the optimal values for $\bm{\theta}_{hyper-s}$ and $\bm{\theta}_{hyper-o}$.}
    \item {$M \doteq F \left( \bm{\theta}_{hyper-o}, \bm{\theta}_{hyper-s} \right), \quad \bm{\theta}_{hyper-o},\; \bm{\theta}_{hyper-s} \in \bm{\Theta}_{hyper}$}
\end{enumerate}
Having a model that makes predictions, we want to evaluate a distance between $\mathbf{\hat{Y}}$ and $\mathbf{Y}$, and thus use a set of metrics $\mathbf{S}$. We define two sets of $\mathbf{S}$---one used to adjust $\bm{\theta}_{learn}$ during model fitting, \mbox{i.e., $\bm{S}_{fit}$}; and one used to evaluate predictions made with $\mathbf{X}_{eval}$, \mbox{i.e., $\mathbf{S}_{eval}$}. $\mathbf{S}_{eval}$ generally includes or extends $\mathbf{S}_{fit}$, but it can be an independent set of evaluation metrics.

\bibliographystyle{ametsocV6}
\bibliography{01.1_references}

\end{document}



\newcommand \theabstract {\clearpage}

\maketitle

\section{Scientific Motivation}

The prediction of tropical cyclogenesis (TCG) is crucial for reducing the risk of damage to human life, ecosystems, and infrastructure \citep[e.g.,][]{pielke2003hurricane,emanuel2005increasing}. However, the fundamental processes behind genesis remain poorly understood, as noted in recent reviews \citep{emanuel2018100, tang2020recent}. While various phenomena, ranging from large-scale events like African Easterly Waves and the Madden-Julian oscillation \citep{klotzbach2014madden} to microphysical processes like ice deposition, affect genesis, the non-linear interactions between these factors make it challenging to predict. Even the role of an initial disturbance in genesis is not well established, as some studies suggest that such disturbances may merely indicate the position and timing at which genesis manifests rather than being necessary from a climatological perspective \citep{patricola2018response}. Smaller-scale environmental processes, including dynamical factors such as low-level relative vorticity, thermodynamic factors such as sea surface temperature, convective activity, and humidity, radiative processes, and microphysical processes such as latent heat release, have all been shown to influence to genesis \citep{yang2020interactive, ruppert2020critical}. Given the variety of scales and processes involved and the lack of physical understanding, predicting tropical cyclogenesis remains challenging and an active research topic.

Existing approaches to predict the occurrence of genesis can be classified into two categories. On the one hand, we can predict whether an existing disturbance (e.g., a tropical cloud cluster) will intensify to become a tropical cyclone \citep[TC,][]{zhang2019prediction, hennon2003forecasting, zhang2021predicting, chang2021machine}. On the other hand, we can estimate the probability of formation of a tropical cyclone in a given time window (e.g., in the following 24h, 48h, or 120h) given a snapshot of atmospheric conditions (e.g., reanalysis data, analysis data, forecast data, and/or satellite imagery) \citep{schumacher2009objective,schumacher2014update}.

In the latter case, genesis forecasting can be framed as a \textit{binary classification} problem, i.e., attributing a value of $0 $ to non-genesis data points and of $1 $ to genesis. However, several challenges set genesis probability estimation at a given location apart from standard classification problems. First, the class imbalance is extreme ($\approx 1:10,000$) as a tropical cyclone forms on average less than $0.1\% $ of the time at a given location in the Tropics. Second, even for simple models, predictions remain difficult to trust because probabilities are not systematically calibrated, and difficult to interpret because explanations are not usually provided alongside statistical predictions.

We aimed to address these limitations in the context of an existing operational product, the Tropical Cyclone Formation Probability Guidance Product (henceforth TCFP), by designing a rule-based algorithm based on transparent threshold to alleviate the class imbalance problem. 

\section{Base Model Details}
\label{sp1:TCFP_details}

To establish a baseline for our study, we developed an implementation of NOAA’s 2014 Tropical Cyclone Formation Probability Guidance Product \citep[TCFP,][]{schumacher2014update,schumacher2009objective}, which relies on the 14 predictors listed in Tab~\ref{tab:predictors} and defined below in section S2a. TCFP employs a three-step method for predicting TCG with a lead time of 24 hours: First, predictors are thresholded such that no more than a predefined percentage of observed events are removed, effectively undersampling the data. Second, the undersampled population is used to train a Linear Discriminant Analysis (``LDA'' hereafter) algorithm whose task is to discriminate between non-events and cyclogenesis events. Finally, the discriminant value is interpolated and scaled to yield a formation probability in each basin that is consistent with the number of observed TC cases in the training set. 

\begin{figure*}
    \centering
    \includegraphics[width=\textwidth]{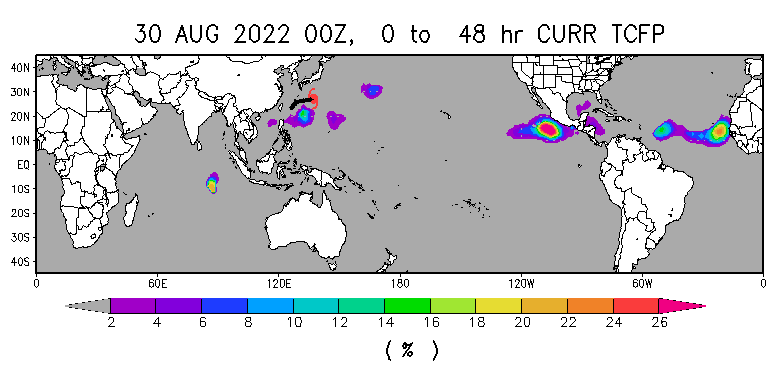}
    \caption{Map of the Aug. 30\textsuperscript{th}, 2022 genesis probability generated by TCFP v3 on Aug. 30. The TCFP map was retrieved from the \href{https://www.ssd.noaa.gov/PS/TROP/TCFP/}{TCFP Website}}
    \label{fig:tcfp_product}
\end{figure*}

Once the algorithm has been trained, it can be used as a prediction product. To do this, the defined filters are applied to forecast fields. The filtered fields are in turn fed one data point at a time into the oceanic basins' corresponding LDA, producing a map similar to Fig~\ref{fig:tcfp_product}, which shows a sample probability plot from the online product. To make sure that we improve upon the operational baseline, we tested our methodology using the TCFP developmental dataset, described in section S2b.

\begin{table}[]
    \centering
    \caption{TCFP relies on 14 dynamical, thermodynamic, and climatological predictors listed below. \label{tab:predictors}}
    \begin{tabular}{c|c|c}
    Predictor name & Unit & Symbol\tabularnewline
    \hline 
    \hline 
    850-200hPa Vertical Shear & $m \; s^{-1}$ & $\Delta_{850-200}U$\tabularnewline
    \hline 
    Weekly Sea Surface Temperature & $^{\circ}C$ & $\mathrm{SST}$\tabularnewline
    \hline 
    850hPa H-Wind Divergence & $ s^{-1}$ & $\delta_{850}$\tabularnewline
    \hline 
    Latitude & $^{\circ}$ & $\lambda$\tabularnewline
    \hline 
    Mean Sea Level Pressure & hPa & $p_{s}$\tabularnewline
    \hline 
    Land fraction & $\%$ & $\mathrm{LF}$\tabularnewline
    \hline 
    850hPa Relative Vorticity  & $ s^{-1}$ & $\zeta_{850}$\tabularnewline
    \hline 
    Cloud Cleared W.V. Bright. Temp. & $^{\circ}C$ & $T_{b}$\tabularnewline
    \hline 
    Clim. Probability of TCG & $\%$ & $p_{\mathrm{clim}}$\tabularnewline
    \hline 
    850hPa Temperature Advection & $^{\circ}C \; s^{-1}$ & $T_{\mathrm{adv}850}$\tabularnewline
    \hline 
    Distance to Existing TC & $km$ & $d_{\mathrm{TC}}$\tabularnewline
    \hline 
    850-200hPa Vertical Instability & $^{\circ}C$ & $\left\langle \Delta\theta_{e}\right\rangle _{850-200} $\tabularnewline
    \hline 
    Percent Cold Pixel Coverage & $\%$ & $\mathrm{PCCOLD}$\tabularnewline
    \hline 
    600hPa Relative Humidity  & $\%$ & $\mathrm{RH}_{600}$\tabularnewline
    \end{tabular}
    
\end{table}


\subsection{List of Predictors}

\noindent \textbf{850 - 250 hPa Vertical Shear} ($\Delta_{850-200}U$) \\ Expressed in $m \; s^{-1}$, the magnitude of the 850 - 250 hPa vertical shear is calculated from the CFSR/GFS winds as 
    \[\Delta_{850-200}U=\sqrt{(u_{850}-u_{200})^2+(v_{850}-v_{200})^2}\\,\]
where \(u_{850}\) \((v_{850})\) and \(u_{200}\) \((v_{850})\) are the zonal (meridional) winds at 850 hPa and 200 hPa, respectively.\\

\noindent \textbf{850 hPa Relative Vorticity} ($\zeta_{850}$) \\ Expressed in $10^{-6} \; s^{-1}$, the 850 hPa relative vorticity is calculated from the CFSR/GFS winds as
\[\zeta_{850}=\frac{\partial v_{850}}{\partial x}-\frac{\partial u_{850}}{\partial y},\]
where \(u_{850}\) and \(v_{850}\) are the zonal and meridional winds at 850 hPa, respectively. \\ 

\noindent \textbf{850 hPa Horizontal Wind Divergence} ($\delta_{850}$) \\ Expressed in $10^{-6} \; s^{-1}$ the 850 hPa horizontal wind divergence is calculated from the CFSR/GFS 850 hPa winds as
\[\delta_{850}=\frac{\partial u_{850}}{\partial x}+\frac{\partial v_{850}}{\partial y},\]
where \(u_{850}\) and \(v_{850}\) are the zonal and meridional winds at 850 hPa, respectively. \\

\noindent \textbf{Mean Sea Level Pressure} ($p_s$) \\ Expressed in hPa, the mean sea level pressure is the CFSR/GFS pressure at sea level.\\

\noindent \textbf{600 hPa Relative Humidity} $\left(RH_{600}\right)$ \\ Expressed in \% of saturated vapor, the 600 hPa relative humidity was obtained directly from the CFSR/GFS data files. \\

\noindent \textbf{850 hPa Temperature Advection} $\left(T_{adv850}\right)$ \\ Expressed in $10^{-6} \; ^\circ C\; s^{-1}$, the 850 hPa temperature advection is calculated from the CFSR/GFS wind and temperature fields by
\[T_{adv850}=u_{850}\frac{\partial T_{850}}{\partial x}+v_{850}\frac{\partial T_{850}}{\partial y},\]
where \(u_{850}\) and \(v_{850}\) are the zonal and meridional winds at 850 hPa, respectively, and \(T_{850}\) is the temperature at 850 hPa. \\ 

\noindent \textbf{850 - 200 hPa Vertical Instability} $\left( \left\langle \Delta \theta_e \right\rangle_{850-200} \; \right)$ \\ Expressed in $^\circ C$, the vertical instability parameter is here defined as the vertically-averaged temperature difference between the equivalent potential temperature of a parcel lifted from the surface to 200 hPa, and the saturation equivalent potential temperature of the environment.\\ 

\noindent \textbf{Weekly Sea Surface Temperature} $\left( SST \right)$ \\ Expressed in $^\circ C$, the weekly sea surface temperature is taken from the Reynolds SST product made available by the National Hurricane Center and NOAA \citep{RSST_page}.\\ 

\noindent \textbf{Percent Cold Pixel Coverage} $\left( PCCOLD \right)$ \\ Expressed in $ \%$, The percent of pixels colder than -40 degree in GOES' 6.7$\mu m$ water vapor channel. All full disk images within 3 hours after and 6 hours before each synoptic time are included, so that this parameter represents the amount of sustained deep convection.\\ 

\noindent \textbf{Cloud Cleared Water Vapor Brightness Temperature} $\left( T_{b} \; \right)$ \\ Expressed in $^\circ C$, this field describes average 6.7$\mu m$ water vapor channel brightness temperature, after the cold pixels in PCCD (defined above) have been eliminated. This parameter is a measure of mid- to upper-level moisture.\\ 

\noindent \textbf{Land Fraction} $\left( LF \right)$ \\ Expressed in $\%$, the proportion of the area within a 500-kilometer radius that is occupied by land.\\ 

\noindent \textbf{Distance to Existing TC} $\left( d_{TC} \right)$ \\ Expressed in $km$ with a range of $5\cdot10^2 \; - \; 10^4 \;km$

\subsection{Developmental Dataset and Training-Test Split}

The TCFP developmental dataset has a 1\textdegree $\;$resolution and spans the full 360\textdegree$\;$of longitude and 45\textdegree$\;$poleward from the equator. It spans year 1995 to 2010, resulting in 1530 TC cases taken from the International Best Track Archive for Climate Stewardship (IBTrACS) and b-decks from the Automated Tropical Cyclone Forecasting System (ATCF). As a pre-processing step, the data were spatially smoothed using a 500km-radius moving average, and temporally smoothed using a 24h window, limiting the fine-scale information we can learn from this dataset. For each case considered, all grid points within a 500km radius of the TC genesis event are marked as a positive case.

\section{Proposed Improvements}
\label{sp2:DDT_details}

\subsection{Data-Driven Filtering \label{sub:Undersampling}}

In this section, we address the class imbalance problem by developing a rule-based, interpretable, targeted filtering algorithm that can be used at inference time. Recent reviews on methods for addressing class imbalance \citep[e.g.,[][]{rendon2020data}) point out the popularity of over- and under-sampling, with well-known drawbacks, e.g., potential for overfitting, underfitting, or deletion of information useful for defining the decision boundary. Unlike oversampling, undersampling makes problems less computationally expensive and has shown to have a positive effect on classifier performance \citep{guzman2021dbig}. As such, we choose two undersampling techniques as baselines: (1) random undersampling, which is useful in domains with a large class imbalance that contain large amounts of data irrelevant to the classification task at hand \citep{japkowicz2002class}; and (2) the undersampling implemented in the original TCFP.

\subsection{Theory}

\begin{figure}
    \centering
    \includegraphics[width=\columnwidth]{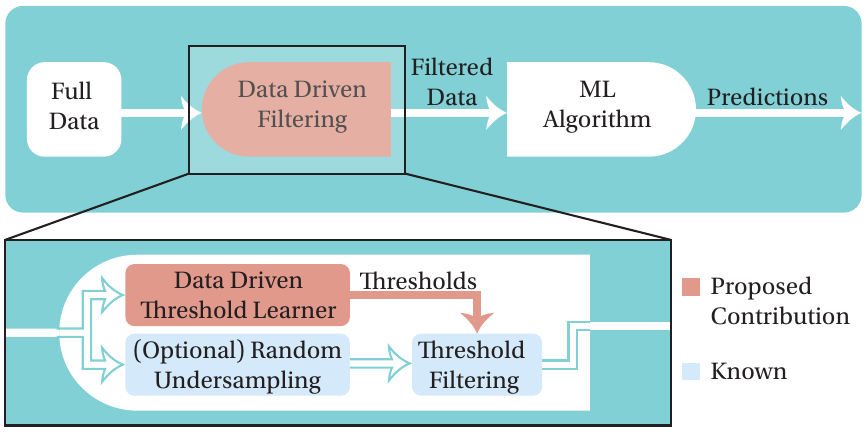}
    \caption{Our proposed Data-Driven Filtering (DDF) algorithm: Interpretable thresholds are learned on the training set to filter the data at training and inference times, alleviating our dataset's high class imbalance and improving the classifier's performance.}
    \label{fig:DDF}
\end{figure}

Consider a classifier $f$ whose goal is to take in a set of predictor values and predict a binary class (i.e., $f:R^{n} \mapsto \{ 0,1 \}$), where the input is given by $\left\{\mathbf{X}_i\right\}_{i \in n}$ and the ground truth as $\mathbf{Y} \in \left\{ 0, 1 \right\}$.  Denoting the original dataset as $\left(\,\mathcal{X},\,\mathcal{Y}\,\right)$, the goal of our proposed undersampling approach is to find a rebalanced dataset $\left(\,X,\, y\,\right)\subset \left(\,\mathcal{X},\,\mathcal{Y}\,\right)$ such that $p(y = 0) < p (\mathcal{Y} = 0)$ and $p(y = 1) \approx p (\mathcal{Y} = 1)$.

To accomplish this task, we define a selection variable $s$ that takes a value of $1$ if the sample is selected and a value of $0$ if the sample is not selected. We then define a set of filter predictors $\left\{F_j\right\}_{j \in n}$ for which we can find threshold values $\tau_{\text{min}} \;\&\; \tau_{\text{max}}$ such that: 
\begin{equation}
    s = \prod_{j \in F}{\left( \mathcal{X}_j > \tau_{min,j}\right)\left( \mathcal{X}_j < \tau_{max,j}\right)}, \quad s \in \left\{ 0, 1 \right\}
\end{equation}
In order to ensure we have found an appropriate set of filter predictors $F$ and threshold values $\tau_j$, we set the objective to minimize a cost function ${\cal L}$ that is a function of the classifier's confusion matrix:
\begin{equation}
    \mathop{min}_{|F|,|\tau_{max}|, |\tau_{min}| } {\cal L} \left( TP, FP, TN, FN \right),
\end{equation}
noting that the confusion matrix is defined by the true positive $TP$, false negative $FN$, false positives $FP$, and true negative $TN$ populations:
\begin{align*}
    \frac{TP}{N_{\mathrm{samples}}} &= p\left(s = 1 | \mathcal{Y} = 1\right),\  &\frac{FN}{N_{\mathrm{samples}}} &= p\left(s = 0 | \mathcal{Y} = 1\right),\\ \\
    \frac{FP}{N_{\mathrm{samples}}} &= p\left(s = 1 | \mathcal{Y} = 0\right), \  &\frac{TN}{N_{\mathrm{samples}}} &= p\left(s = 0 | \mathcal{Y} = 0\right).
\end{align*} 

\begin{figure}
    \centering
    \includegraphics[width = \columnwidth]{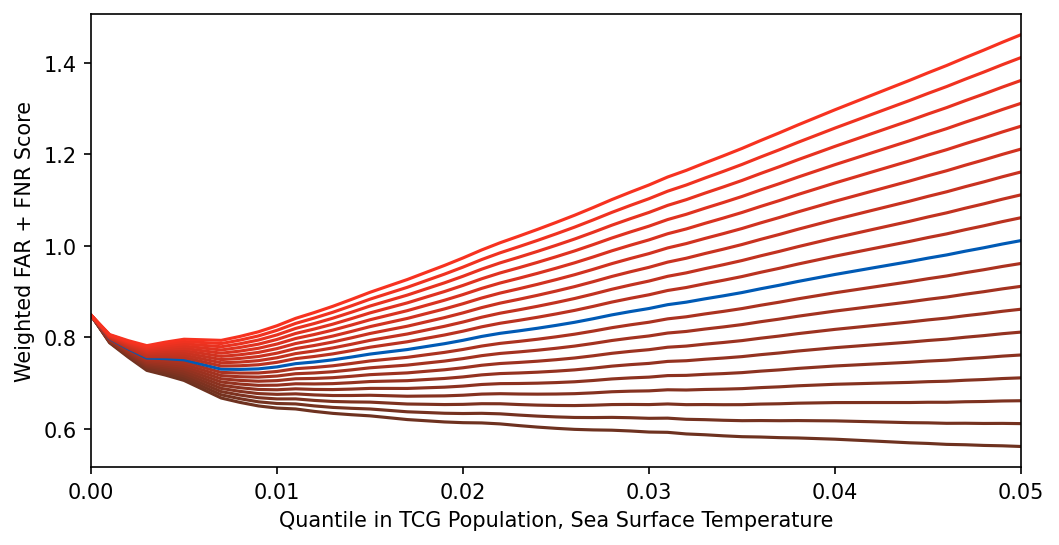}
    \caption[Weighted $FAR + FNR$ score]{Effect of the weighting parameter $\lambda$ on the Weighted $FAR + FNR$ Score. Lines plotted for $\lambda \in \left\{ 1..19\right\}$, $\lambda = 10$ is highlighted in blue.}
    \label{fig:importance_score}
\end{figure}

We wish to affect the minority class (e.g., genesis) as little as possible with the filtering step, attempting to meet the constraint that $p(y = 1) \approx p (\mathcal{Y} = 1)$. By splitting the populations into the positive and negative class, assuming the overall sample size is enough to capture the general behavior of the predictors for both classes, the quantiles of $\left\{ X_i \right \}_{i \in F} | \mathcal{Y} = 1$ are simply the \textit{False Negative Rate} ($FNR$), which is calculated as $FNR = \frac{FN}{ TP + FN }$. The values of the thresholds $\tau_{\text{min}}, \tau_{\text{max}}$ can therefore be derived from the predictor quantiles associated with $s=1$ and the $FNR$ is given a priori by the quantile.\\
The next aspect that must be addressed is the reduction of the majority class (i.e., non-genesis points), with the objective that $p(y = 0) \ll p (\mathcal{Y} = 0)$. The effectiveness of $s$ in removing non-genesis cases can be expressed as the \textit{False Alarm Rate} ($FAR$), which is calculated as $FAR = \frac{FP}{ FP + TN }$. Unlike the $FNR$, the $FAR$ is not known a priori since the $FAR$ of the negative class is not related to the quantile value of the positive class.
The cost function ${\cal L}$ must therefore balance the $FNR$ and $FAR$. In order to do this, we define a hyperparameter $\lambda$ that weights the relative importance of the $FNR$ compared to the $FAR$. The two metrics are then combined in a \textit{Weighted FAR + FNR} score, given by:
\begin{equation}
    \text{Weighted}\; FAR + FNR = \lambda \times \; {FNR} + FAR
\end{equation}
$\lambda$ shifts the location of the minimum value of the Weighted $FAR + FNR$ score, as shown in Fig~\ref{fig:importance_score}.

Once the predictor and threshold combination with the lowest \mbox{weighted $FNR + FAR$} score is found, the dataset is filtered using the threshold. The analysis is repeated with the remaining predictors, and thresholds are found in this way until additional predictors exhibit no significant additional discriminative capacity (e.g., when the lowest score is $\geq0.98$). An example is given in Tab~\ref{tab:importance}.

\begin{table*}[]
    \centering
    \small
    \caption[Sequential Thresholding Scores]{Weighted $FAR + FNR$ scores for sequentially selected thresholds in oceanic basins}
    \setlength{\tabcolsep}{6pt}
    \begin{tabular}{c c c c c c c}
         &\multicolumn{2}{c}{North Atlantic} & \multicolumn{2}{c}{Northeast Pacific} & \multicolumn{2}{c}{Northwest Pacific}  \tabularnewline
         &Predictor & Score & Predictor & Score & Predictor & Score \tabularnewline
        \hline  \tabularnewline
         1\textsuperscript{st} Pass & RSST & $0.7353 \pm 6\mathrm{e}{-4}$& RSST & $0.6325 \pm 1\mathrm{e}{-3}$& RSST & $0.5844 \pm 6\mathrm{e}{-4}$ \tabularnewline
         2\textsuperscript{nd} Pass & PCCD & $0.8335 \pm 6\mathrm{e}{-4}$& PCCD & $0.8325 \pm 8\mathrm{e}{-3}$& PCCD & $0.6375 \pm 7\mathrm{e}{-4}$ \tabularnewline
         3\textsuperscript{rd} Pass & THDV & $0.9242 \pm 5\mathrm{e}{-4}$& - & - & VSHD & $0.8518 \pm 8\mathrm{e}{-4}$
         
    \end{tabular}
    \label{tab:importance}
\end{table*}

\subsection{Inconclusive Results}

\subsubsection{Methods}
For random undersampling, during training, we select samples from the majority class in the training dataset at random while retaining the minority class in its entirety. This introduces a sample selection bias and requires model recalibration after training. For the TCFP baseline, the datapoints are screened using thresholds selected by a domain expert after conducting a per ocean basin examination of the predictor fields \citep{schumacher2009objective}. The thresholds are applied over all predictor fields, constrained only in that the thresholds must not eliminate more than a given percentage of the genesis cases in the training dataset. To keep the interpretable nature of the original TCFP filtering approach while improving upon its semi-empirical thresholds, we use our DDF strategy and leverage the labeled nature of the dataset to transparently produce threshold values for the predictors (illustrated in Fig~\ref{fig:DDF}). In short, DDF calculates the predictor quantiles associated with the observed events in the dataset and uses the distributions' tail ends to find appropriate thresholds for each predictor. By comparing the miss rate and false alarm rate after filtering, the performance of each predictor as a bounding variable is quantified, which allows us to only threshold on predictors significantly improving discriminative skill. We can then sequentially select the bounding variables until no significant skill gain is observed.

\subsubsection{Findings}
While we obtained improvements by using the DDF strategy, measured by an increase in the Mathews Correlation Coefficient (MCC) (as described by \cite{MCC}) and less sensitivity to the data folds, our results were inconclusive. This was in part due to the fact that while we observed an improvement in MCC, the order of improvement was much less than the observed variance of the score across folds. This challenge motivated the creation of a roadmap to better navigate the complexities of model replication and performance assessment.

\section{Questionnaire}
\label{sp3:Questionnaire}
Questionnaires are meant to be answered with a value ranging from 0 (Not Equivalent/Not Applicable) to 3 (Fully Equivalent). The question categories are Dataset (DS), Metrics (S), and Model (M).

\begin{table*}
    \centering
    \caption{Equivalence Comparison Questions for Fitting Axis}
    \begin{tabular}{|c|c|l|l|}
    \hline
    \textbf{No.} & \textbf{Type} & \textbf{Question} & \textbf{Equivalence (1-4)} \\
    \hline \hline
    1 & ds & Is the computational processing of the fitting dataset equivalent? & \\
    \hline
    2 & ds & Are the predictors selected for model fitting equivalent? & \\
    \hline
    3 & ds & Is the method used to select the predictors used for fitting equivalent? & \\
    \hline
    4 & ds & Is the definition of the target event equivalent? & \\
    \hline
    5 & ds & Are the preprocessing steps applied to the predictors before fitting equivalent? & \\
    \hline
    6 & ds & Are the preprocessing steps applied to the target variable(s) before fitting equivalent? & \\
    \hline
    7 & ds & Is the dataset size used for model fitting equivalent? & \\
    \hline
    8 & ds & Is the method for partitioning the dataset into folds for use in fitting equivalent? & \\
    \hline
    9 & S & Is the way the fitting metrics are calculated and/or implemented equivalent? & \\
    \hline
    10 & S & Is the set of metrics chosen for use in model fitting equivalent? & \\
    \hline
    11& S  & Is the use of the chosen metrics within fitting optimization equivalent? & \\
    \hline
    12 & S & Are the observed values of the metrics chosen for fitting equivalent? & \\
    \hline
    13 & M & Is the computational implementation of the algorithm equivalent? & \\
    \hline
    14 & M & Is the range of hyperparameters explored equivalent? & \\
    \hline
    15 & M & Is the strategy used to search the hyperparameter space equivalent? & \\
    \hline
    16 & M & Are the best hyperparameters discovered during the search equivalent? & \\
    \hline
    17 & M & Are the learnable parameters observed after fitting equivalent? & \\
    \hline
    \end{tabular}
    \label{tab:equivalence_numbered}
\end{table*}

\begin{table*}[]
    \centering
    \caption{Equivalence Comparison for Position in Inference Axis}
    \begin{tabular}{|c|c|l|l|}
    \hline
    \textbf{No.} & \textbf{Type} & \textbf{Question} & \textbf{Equivalence (1-4)} \\
    \hline \hline
    1 & ds & Is the computational processing of the evaluation dataset equivalent? & \\
    \hline
    2 & ds & Are the predictors used with the model equivalent? & \\
    \hline
    3 & ds & Is the definition of the target event equivalent? & \\
    \hline
    4 & ds & Are the preprocessing steps applied to the predictors before evaluation equivalent? & \\
    \hline
    5 & ds & Are the preprocessing steps applied to the target variable(s) before evaluation equivalent? & \\
    \hline
    6& ds  & Do the model fitting and model evaluation datasets contain equivalent information? & \\
    \hline
    7 & ds & Is the dataset size used for model evaluation equivalent? & \\
    \hline
    8 & ds & Is the method for partitioning the dataset into folds for evaluation equivalent? & \\
    \hline
    9 & S & Is the way the evaluation metrics are calculated and/or implemented equivalent? & \\
    \hline
    10 & S & Is the set of metrics chosen for model evaluation equivalent? & \\
    \hline
    11 & S & Are the observed values of the metrics chosen for fitting equivalent? & \\
    \hline
    12 & M & Is the computational implementation of the algorithm equivalent? & \\
    \hline
    13 & M & Are the structural hyperparameters of the model equivalent? & \\
    \hline
    14 & M & Are the learned parameters in the model equivalent? & \\
    \hline
    \end{tabular}
    \label{tab:evaluation_equivalence_numbered}
\end{table*}

\clearpage
\bibliographystyle{ametsocV6}
\bibliography{02.1_references}